\documentclass[conference]{IEEEtran}
\IEEEoverridecommandlockouts
\usepackage{cite}
\usepackage{amsmath,amssymb,amsfonts}
\usepackage{algorithmic}
\usepackage{graphicx}
\usepackage{textcomp}
\usepackage{xcolor}
\usepackage{url}
\usepackage{bbm}
\usepackage{booktabs}
\usepackage{mathtools}
\usepackage{wrapfig}
\usepackage[normalem]{ulem}
\usepackage{mathrsfs}
\usepackage{multirow}
\usepackage{pifont}
\usepackage{subfigure}
\usepackage{CJKutf8}
\usepackage[linesnumbered,ruled,vlined]{algorithm2e}

\newtheorem{definition}{Definition}

\def\model{HiPSTG}

\def\BibTeX{{\rm B\kern-.05em{\sc i\kern-.025em b}\kern-.08em
    T\kern-.1667em\lower.7ex\hbox{E}\kern-.125emX}}
\begin{document}

\title{Resolving the Imbalance Issue in Hierarchical Disciplinary Topic Inference via LLM-based Data Augmentation}

\author{\IEEEauthorblockN{Xunxin Cai$^{1,2}$, Meng Xiao$^{1,2, *}$\thanks{$*$ Corresponding Author.}, Zhiyuan Ning$^{1,2}$, Yuanchun Zhou$^{1,2}$}
\IEEEauthorblockA{\textit{$^1$Computer Network Information Center, Chinese Academy of Sciences, China} \\
\textit{$^2$University of Chinese Academy of Sciences, China}} \\
\{xxcai, shaow, ningzhiyuan, zyc\}@cnic.cn
}

\maketitle

\begin{abstract}
In addressing the imbalanced issue of data within the realm of Natural Language Processing, text data augmentation methods have emerged as pivotal solutions. 
This data imbalance is prevalent in the research proposals submitted during the funding application process. 
Such imbalances, resulting from the varying popularity of disciplines or the emergence of interdisciplinary studies, significantly impede the precision of downstream topic models that deduce the affiliated disciplines of these proposals. 
At the data level, proposals penned by experts and scientists are inherently complex technological texts, replete with intricate terminologies, which augmenting such specialized text data poses unique challenges. 
At the system level, this, in turn, compromises the fairness of AI-assisted reviewer assignment systems, which raises a spotlight on solving this issue. 
This study leverages large language models (Llama V1) as data generators to augment research proposals categorized within intricate disciplinary hierarchies, aiming to rectify data imbalances and enhance the equity of expert assignments. 
We first sample within the hierarchical structure to find the under-represented class.
Then we designed a prompt for keyword-based research proposal generation.
Our experiments attests to the efficacy of the generated data, demonstrating that research proposals produced using the prompts can effectively address the aforementioned issues and generate high quality scientific text data, thus help the model overcome the imbalanced issue.
\end{abstract}


\section{Introduction}

Inference of the disciplinary topic for a given research proposal is a preliminary step for automating the peer-review system~\cite{xiao2023hierarchical}, in which an accurate discipline code could help the funding administrator to assign domain-related experts, thus conducting a fair evaluation. 
Due to the hierarchical nature of disciplines, this type of topic inference task can be defined as a hierarchical multi-label classification task~\cite{xiao2021expert,xiao2022should,ye2023needed}. 
However, due to the different developments, plans, and divisions of major disciplines (e.g., Information Science and Mathematical Science, etc.), there is an imbalance in the number of applications related to these hierarchical discipline labels. 
This data-level imbalance can further affect the accuracy of automated topic inference models for some more minor category disciplines. 
Further, this may result in some novel disciplines being reviewed by experts who are irrelevant to the field, further constraining the development of emerging disciplines~\cite{xiao2023fair}. 
Artificial Intelligence (AI) has made a significant approach in related research areas, e.g., domain knowledge extraction and construction~\cite{zhou2020survey}, scholar data mining~\cite{qiao2022rpt}. 
Recently, the Data-Centric AI method has become a new research trend, which aims to select the critical part of the given data via reinforcement learning~\cite{xiao2023beyond}, or transform the provided feature to an optimal state~\cite{xiao2023traceable,xiao2023traceable2,wang2023reinforcementenhanced}.

Firstly, existing strategies for addressing text classification imbalances can be broadly categorized into three primary paradigms: (1) Methods at the data level focus on rectifying class distribution disparities through techniques like down-sampling and over-sampling~\cite{mullick2019generative,kim2019imbalanced}. (2) In contrast, algorithm-level techniques prioritize the enhancement of minority classes by incorporating specific penalty functions~\cite{ling2008cost,puthiya2014optimizing}. (3) Hybrid methods adeptly fuse the strengths of both sampling and algorithmic strategies to combat imbalances~\cite{krawczyk2016learning}. 
However, these methodologies are predominantly tailored to address classification from a flat perspective. 
When applied to the domain of disciplinary topic inference, the inherently hierarchical nature of discipline structures poses significant challenges for these traditional imbalance learning approaches. 
Moreover, these methods often overlook the rich semantic and structural intricacies embedded within the discipline hierarchies. 
This raises a crucial question: \textbf{Is there potential in leveraging this long-overlooked information to devise robust data augmentation strategies?}

Secondly, in the contemporary landscape of artificial intelligence, large language models (LLMs) like GPT-3~\cite{gpt3}, GPT-4, ChatGPT~\cite{openai2023gpt4}, and Llama~\cite{llama} have emerged as trailblazers in the domain of text generation. 
The capabilities they usher in span a wide array of applications and have been instrumental in generating high-quality textual content~\cite{uchendu2021turingbench}, expeditiously retrieving knowledge~\cite{ng2023simplyretrieve}, and even in the realms of code generation and optimization~\cite{zhong2023study}. 
Those LLMs are also adopted as data augmentation methods for resolving data imbalance issue~\cite{dai2023auggpt}, or data augmentation for hard-collector dataset~\cite{guan2023cohortgpt} (e.g., clinical records).
However, those approaches lack the perspectives of hierarchical label structure.
Given their multifaceted prowess, one is naturally inclined to ponder: \textbf{What enhancements can these formidable LLMs introduce to imbalanced scientific text data?} The potential for these models to revolutionize traditional methodologies and introduce novel paradigms is an exciting avenue of exploration.

To answer those questions, we proposed a Hierarchical-disciplinary-structure Prompted Scientific Text Generator (\textbf{HiPSTG}) to enhance the imbalanced hierarchical organized dataset. 
In detail, HiPSTG first samples the minority class from the disciplines. 
Then, the method constructs a prompt sentence using semantic and structural information from the hierarchical discipline structure. 
Finally, the constructed prompt will be sent to the large language model, while the generated text will be supplied to the training set.
To summarize, the main contributions of this paper are listed as follows:

\begin{itemize}
    \item[(1)] We extend the application of a large language model into the data augmentation for imbalanced hierarchical scientific text classification.
    \item[(2)] The proposed method can utilize the semantic and structural information from the given hierarchical discipline structure to construct an insightful prompt, thus generating unique and reasonable records.
    \item[(3)] Our experiments show the potential of the proposed methods for both imbalance learning and data augmentation on hierarchical discipline structure. 
\end{itemize}





\section{Topic Inference Problem in Research Proposal}

\begin{definition}[\textbf{Research Proposal}]
When scholars or researchers seek funding, they craft grant applications which are essentially structured compositions containing several elements, notably the \textit{Title}, \textit{Abstract}, \textit{Keywords}, and \textit{Research Areas}. We represent a grant application proposal with \( D \). The various components within this proposal are represented as \( D=\{d_t\}_{t=1}^{\mid T\mid } \), where each document type is signified by \( t \) belonging to the set \( T \) of all document classifications. The count of these types is \( \mid T\mid \), and \( d_i \) signifies the document of type \( t_i \). Documents are essentially sequences of word tokens, which we symbolize as \( d_i = [w^{1}_i,w^{2}_i,...,w_{i}^{\mid d_i\mid }] \), where \( w^{k}_i \) is the kth word token within document \( d_i \).
\end{definition}

\smallskip

\begin{definition}[\textbf{Hierarchical Discipline Structure}]
The hierarchical disciplinary structure, symbolized by \( \gamma \), can either be a DAG or a tree structure. It's made up of academic domains and includes a directed \textit{Belongs-to} association pointing from a domain to its sub-domains. The nodes within this discipline, set \( C=\{C_0\cup C_1\cup ...\cup C_H\} \), are systematically arranged across \( H \) hierarchical stages. Here, \( H \) defines the depth of hierarchy, and \( C_k=\{c^i_k\}^{\mid C_k\mid }_{i=1} \) stands for the set of disciplines at the $i$-th level. The root level is represented by \( C_0=\{root\} \). To articulate the interconnections among different disciplines, we introduce \( \prec \) - a partial order showcasing the \textit{Belongs-to}. $\prec$ is asymmetric, anti-reflexive and transitive\cite{wu2005learning}:
\begin{small}
 \begin{align} 
   & \bullet \text{The only one greatest category }\textit{root}\text{ is the root of the } \gamma,  \nonumber\\
   & \bullet \forall c^x_i \in C_i, c^y_j \in C_j, c^x_i \prec c^y_j \to \space c^y_j \not\prec c^x_i,  \nonumber\\
   & \bullet \forall c^x_i \in C_i, c^x_i \not\prec c^x_i, \nonumber\\
   & \bullet \forall c^x_i \in C_i, c^y_j \in C_j, c^z_k \in C_k, c^x_i \prec c^y_j \land c^y_j \prec c^z_k \to c^x_i \prec c^z_k.  \nonumber
   \nonumber
 \end{align}
\end{small} 

We then express the Structured Academic Discipline \( \gamma \) as a partially ordered set \( \gamma=(C,\prec) \).
\end{definition}

\smallskip

\begin{definition}[\textbf{Disciplinary Topic Inference}]
The task of Disciplinary Topic Inference is approached using a Hierarchical Multi-label Classification framework. Here, a sequence of domain-level-specific label sets symbolizes the grant application's academic codes. This is represented as \( L = [l_0, l_1, l_2,..., l_{H_A}] \), where \( l_0 = \{l_{root}\} \). The prediction process is broken down in a top-down manner, starting from the initial level to a particular level within the academic discipline structure \( \gamma \). The prediction at the k-th level is perceived as a multi-label classification, formulated as:
\[
\Omega(D,L_{<k},\gamma;\Theta) \to L_{k}
\]
Lastly, the likelihood of the label set sequence assignment for the proposal during prediction is expressed as:
\[
P(L\vert D,\gamma;\Theta)=\prod_{k=1}^{H_A} P(l_k\vert D,L_{<k},\gamma; \Theta)
\]
In the training phase, our objective, given the actual labels, is to optimize the above equation.
\end{definition}


\section{Methodology}

\begin{figure*}[!h]
    \centering
    \includegraphics[width=0.9\textwidth]{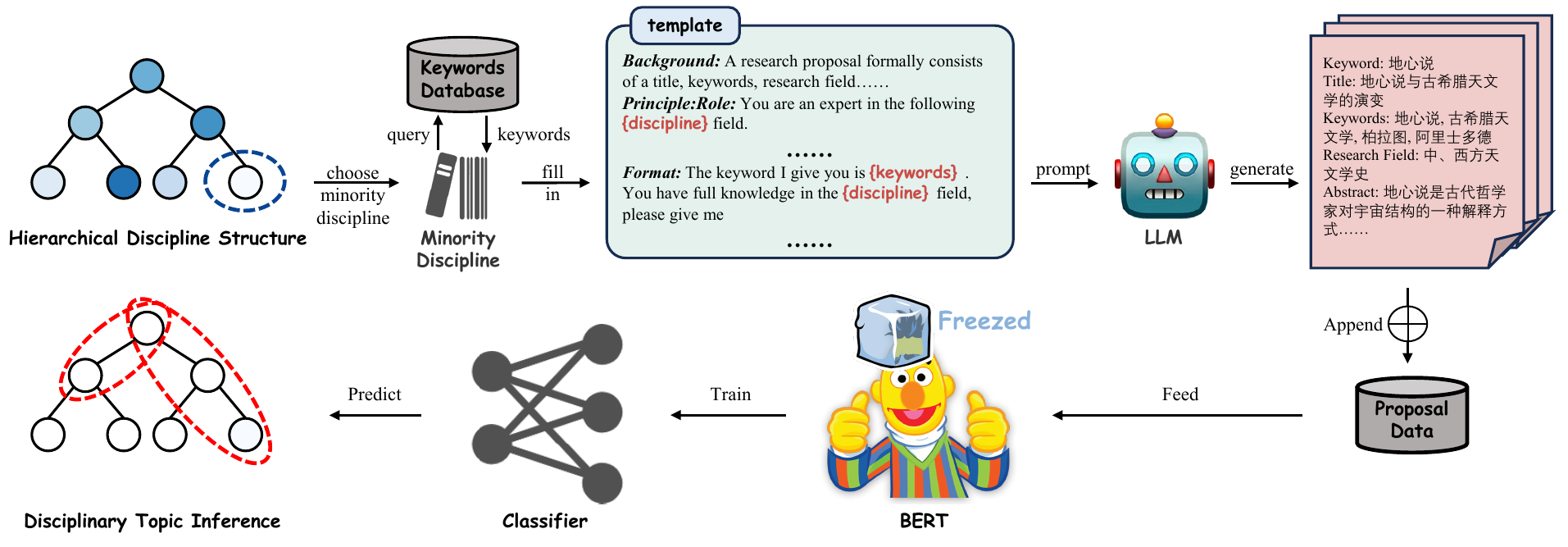}
    \caption{An illustration of the framework of \model .}
    \label{main_fig}
\end{figure*}

Figure~\ref{main_fig} illustrates the overview of the proposed method, \model, which consists of two parts. 
The first part, Structural-informed Minority Class Sampling, aims to obtain the minority class from the hierarchical discipline structure. 
The second part, Prompt Construction, is designed to build the prompt for LLMs-based generation. 
The constructed prompt will then feed into the LLM and the scientific text data will be generated. Overall, given dataset $D$ and their related


\subsection{Structural-informed Minority Class Sampling.} 

Handling imbalanced datasets is crucial in many applications, as having a skewed distribution of classes can lead to models that are biased toward the majority class.
Data augmentation, especially for the minority class, is one way to address this. 
The first stage of HiPSTG is designed to find the minority class that needs to be augmented. 
Algorithm~\ref{algo_disjdecomp} present the minority class sampling method on the hierarchical discipline structure.

Supposed we have the overall training set, denoted as $\{\textbf{D}, \textbf{L}\}$, where $\textbf{D}$ is the set of research proposals and  $\textbf{L}$ is the corresponding label set. 
The discipline set is also provided as $C$. 
For each $c \in C$, we can obtain the number $n_c$ of the related proposal by iteratively scanning the label set $\textbf{L}$. 
We then calculate the minority score of each discipline via:
\begin{equation}
    s_c = \frac{e^{-\frac{n_c}{\lambda}}}{\sum_{i\in C} e^{-\frac{n_i}{\lambda}}}, 
\end{equation}
where the $\lambda$ is a hyperparameter to adjust the focal point, when $\lambda$ is higher, the model will focus more on the classes with fewer samples, and vice versa. $s_c$ is the minority score that can represent the proportion of the generation number.
Then, the generation number of each discipline $c$ can be calculated via: $f_c = T \dot s_c$, where $T$ is the total generation number and $f_c$ denoted the generation number of discipline $c$.

\SetAlgoVlined
\IncMargin{1em}
\setlength{\algomargin}{2em}
\SetCustomAlgoRuledWidth{25em}

\begin{algorithm}
\caption{\textbf{Minority Class Sampling.}}
\label{algo_disjdecomp} 
\SetKwData{Left}{left}\SetKwData{This}{this}\SetKwData{Up}{up}
\SetKwFunction{push}{push}
\SetKw{Or}{||} 
\SetKw{And}{\&\&} 
\SetKw{is}{is}
\SetKw{not}{not}
\SetKw{bk}{Break}
\SetKwData{lb}{left bracket}
\SetKwData{fn}{feature name}
\SetKwData{rb}{right bracket}
\SetKwData{op}{operation}
\SetKwData{sos}{<SOS>}
\SetKwData{eos}{<EOS>}
\SetKwData{sep}{<SEP>}
\SetKw{true}{True}
\SetKwInOut{Input}{input}\SetKwInOut{Output}{output}
\SetKwIF{If}{ElseIf}{Else}{if}{:}{elif}{else:}{}%
\Input{Hierarchical Discipline Structure $\gamma$ Research Proposal Dataset $\{\textbf{D}, \textbf{L}\}$ }
\Output{Sampling probability of each discipline}
\BlankLine 
    \For{$\textbf{each }  nodeId \in \{\textbf{D}, \textbf{L}\}$}{
        $node \leftarrow \gamma.getNode(nodeId[0])$
    
        $node.num \leftarrow node.num + 1$

        $i \leftarrow 1$

        \While{$i < (len(nodeId) + 1) / 2$}{
            $node \leftarrow \gamma.getNode(nodeId[0, i * 2 + 1])$
    
            $node.num \leftarrow node.num + 1$

            $i \leftarrow i + 1$
        }
    }
    $node \leftarrow \gamma.root$

    $Count \leftarrow node.num$

    $S.push(node)$

    $level \leftarrow 1$

    \While{$S \text{ \is \not  } \text{ empty}$}{
        $len \leftarrow S.size()$

        \While{$len \text{ \is \not 0 } $}{
            $node \leftarrow S.pop()$
            
            \uIf{$level \text{ \is 1} $}{
                $node.globalFreq \leftarrow 1.0$
            }
            \uElse{
                $node.globalFreq \leftarrow node.num/Count$
            }

            \For{$\textbf{each } child \in node.children$}{
            
                \uIf{$child \text{ \is \not null}$}{
                
                    $S.push(child)$
                    
                }
                
            }

            $len \leftarrow len - 1$
            
        }
        
    }

    \Return $\gamma;$
\end{algorithm}


\begin{figure*}[!ht]
    \centering
    \includegraphics[width=0.85\textwidth]{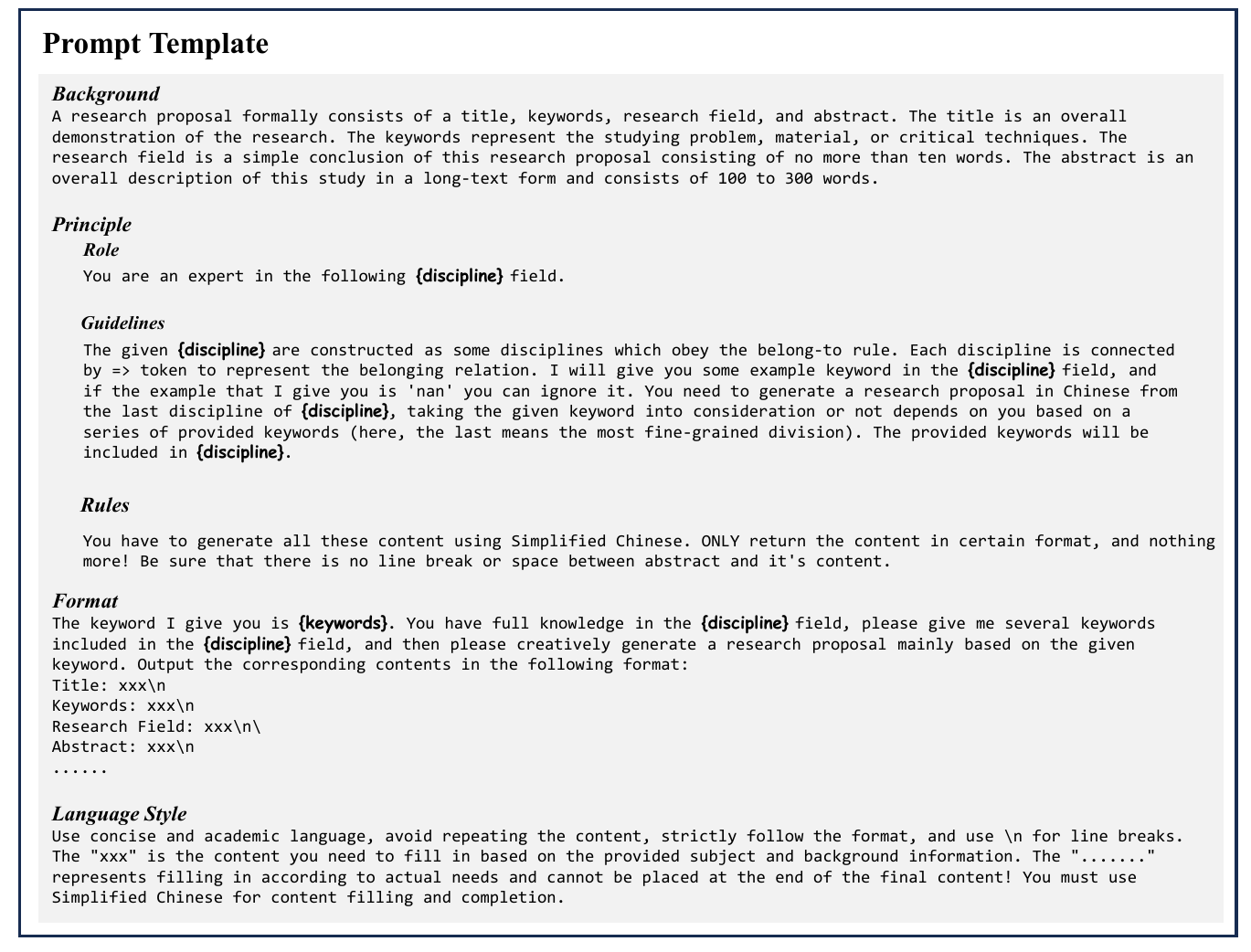}
    \caption{An illustration of the designed prompt.}
    \label{prompt_fig}
\end{figure*}


\smallskip
\subsection{Prompt Design for Scientific Text Generation.}
Technical texts are a type of long and short text composed of keywords related to a specific field. 
In order to effectively generate this type of data and enhance the diversity of the generated data, we interact with the large language model using the Prompt Engineering~\cite{oppenlaender2022prompt} method. 
As shown in Figure~\ref{prompt_fig}, such a 'Prompt' consists of four main parts, namely \textit{Background}, \textit{Principle}, \textit{Format}, and \textit{Language Style}. 
In the Background section, we provide the specific abstract content described in each part of the application and the related length. 
In the Principle section, we define the role of the language model we are talking to, the guiding content (i.e., prompting what it needs to complete), and some formal restrictions on the generated text. 
In the Format section, we strictly constrain the format of the generated text through the Prompt. 
In the Language Style section, we strictly constrain the tone and language type generated by the language model.
In order to prevent the language model from generating meaningless research plans too randomly, we adjust the model's focus by providing two parameters: keywords and disciplines. 
Each selected keyword and discipline is manually provided by experts from funding agencies to guarantee validity.
In this study, we adopted Llama V1 as the backbone LLMs. 
After the pseudo-research proposals are generated, we append them into the training set for the model training.


\section{Experiment}
\begin{figure*}[!t]
\centering
\subfigure[Augmented with 1000 Samples]{
\includegraphics[width=4.4cm]{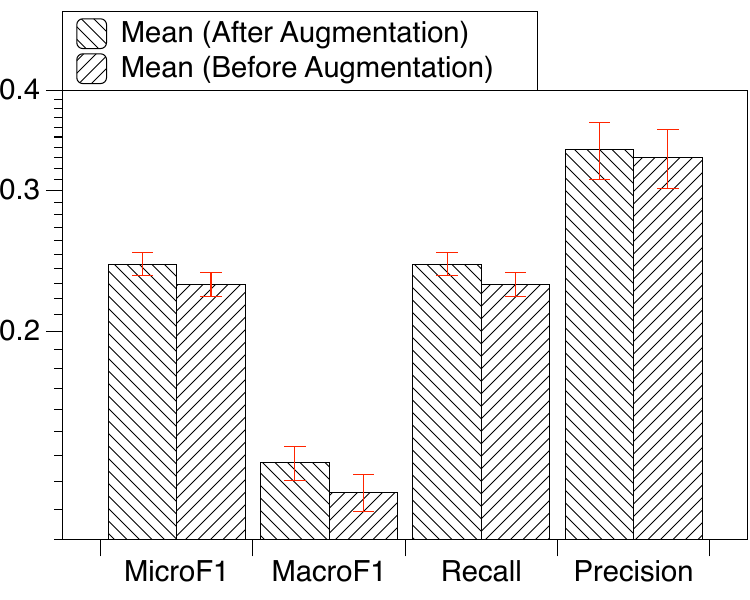}
}
\subfigure[Augmented with 350 Samples]{
\includegraphics[width=4.4cm]{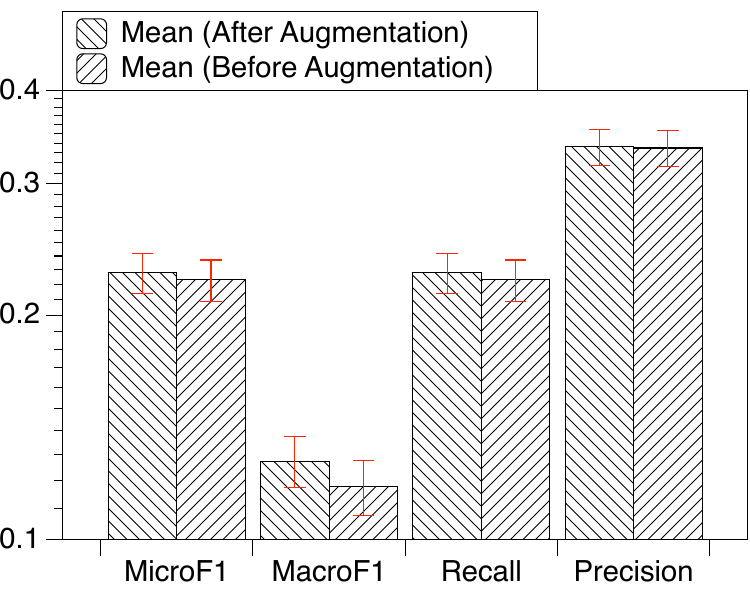}
}
\subfigure[Improvement Comparison]{ 
\includegraphics[width=4.4cm]{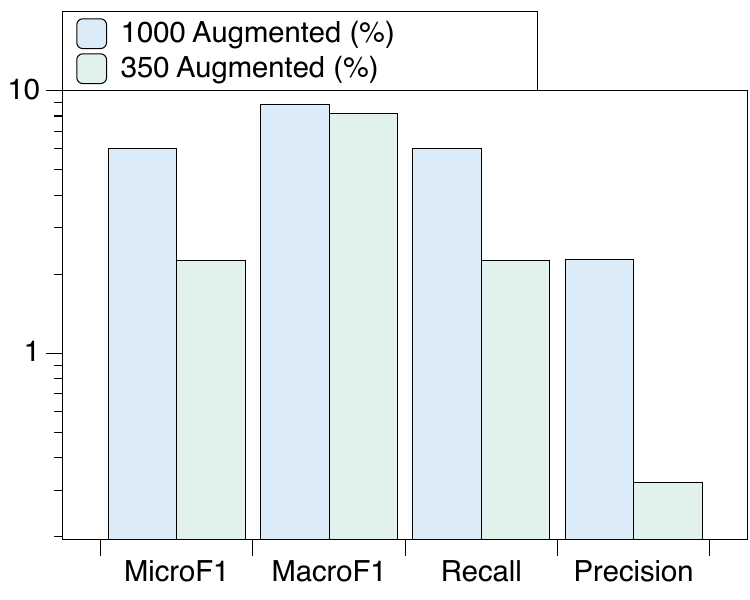}
}
\caption{The illustration of main experimental results. (a-b) The influence of \model\ in different augment number settings. (c) the improvement of two settings.}
\label{main_exp}
\end{figure*}

\subsection{Main Comparison}
From Figure~\ref{main_exp}, we can observe that augmenting the dataset using a large language model showed promising results, especially with 1000 synthetic samples. 
The model's overall performance improved, as evident from the increase in MicroF1, MacroF1, and Recall scores, and Precision scores.
In Figure~\ref{main_exp} (c), when comparing the two augmentation sizes, 1000 samples generally yielded better results by + 8\% in terms of MicroF1. The slight performance dip with 350 samples (+ 6.1\% in terms of MicroF1) suggests that the amount of synthetic data added plays a crucial role in determining the model's success. However, the increased precision with 350 samples is noteworthy, implying fewer false positives.
In summary, using a large language model for data augmentation to address class imbalance shows potential. The choice of the number of samples for augmentation should be made carefully, considering the model's performance-efficiency trade-off. Future work could explore optimizing the number of synthetic samples and refining the generation process to further enhance model performance.


\begin{figure*}[!h]
    \centering
    \includegraphics[width=0.85\textwidth]{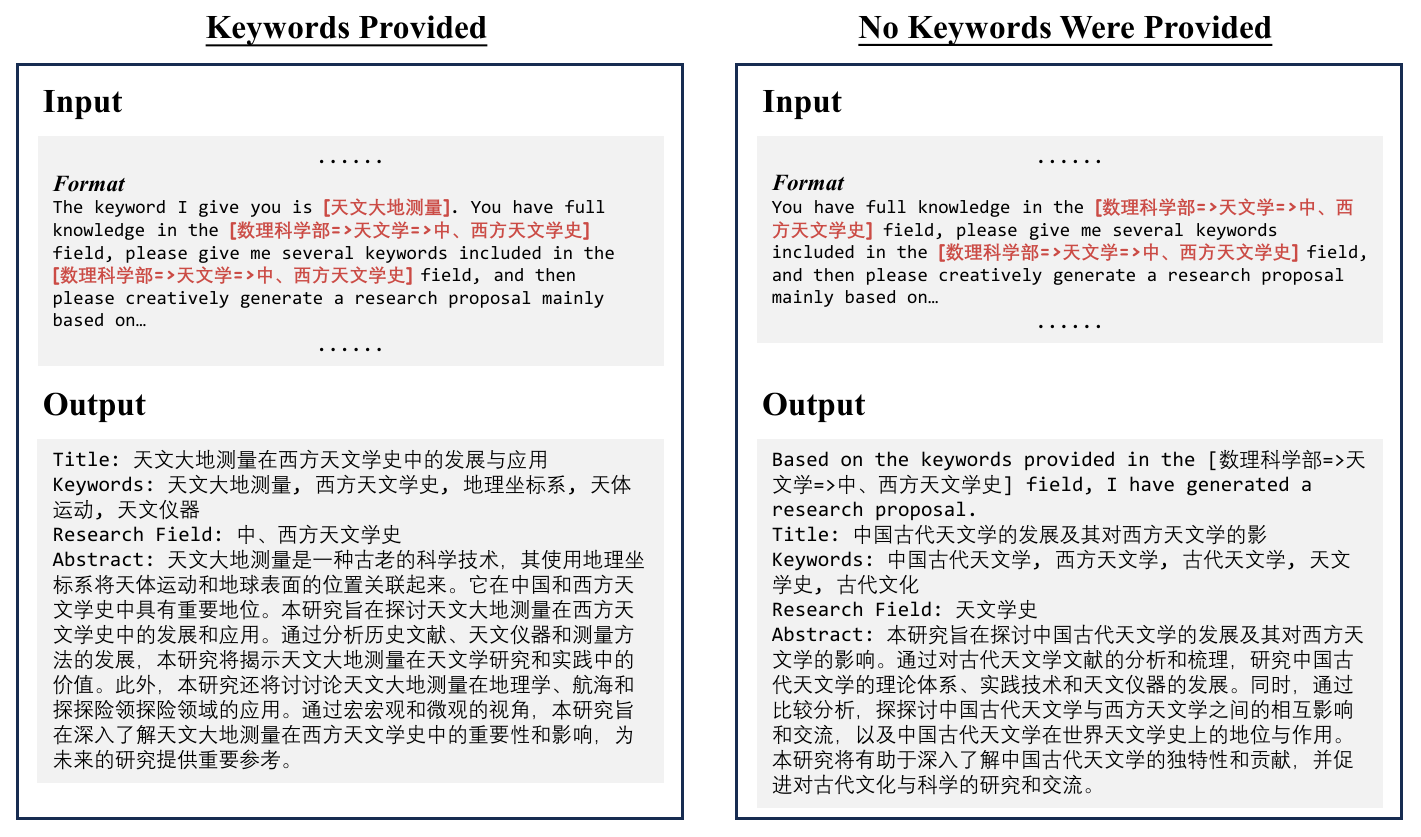}
    \caption{The case study on the generation quality by given or not-given Keywords.}
    \label{promopt}
\end{figure*}

\subsection{Case Study on Generation Strategy}
In Figure~\ref{promopt}, we showcase research proposals generated by Llama using two different strategies: by providing keywords ("Keyword Provided") and without giving keywords ("No Keyword Were Provided"). From the figure, it can be seen that on the left side, we provided the large language model with the keyword "Astronomical Earth Survey". This keyword helps constrain the generated text and offers more semantic diversity and structural diversity in the responses through the provision of multiple keywords. By comparing the generated texts on the left and right, we can observe that by providing a keyword, the large language model can produce more diverse sentence structures and demonstrate superior emergent capabilities, even though some parts of the text might have lexical errors. In contrast, the generated text on the right tends to repetitively focus on the provided sub-disciplines, namely the history of astronomy in both the East and West. This case study validates the effectiveness of our strategy and provides insights for future research.

\section{Conclusion}
This study demonstrates the potential of large language models as data augmenters to address class imbalance in specialized text datasets. By selectively generating high-quality proposals for underrepresented disciplines, the authors were able to improve the fairness and effectiveness of downstream AI systems for expert reviewer assignment. The proposed methods provide an impactful solution to data scarcity in niche domains while ensuring representational equity across the disciplinary hierarchy. Overall, this research highlights the synergistic combination of hierarchy-aware sampling and controllable text generation as a promising approach to alleviating data imbalances in real-world NLP tasks. The novel application to research proposal augmentation opens exciting possibilities for enhancing fairness in AI systems that utilize expert knowledge.

\section{Future Work}
While this work demonstrates promising results in alleviating data imbalance for research proposal augmentation, there remain several worthwhile directions for future investigation. 
One area of focus could be exploring additional large language models and prompt engineering techniques to further enhance the quality and diversity of generated proposals. As new models emerge that are optimized for long-form scientific text generation, leveraging these advances could improve proposal generation fidelity. Secondly, the prompts could be expanded to incorporate other metadata like author details to better emulate real proposals.
Overall, this work opens up an exciting new capability for alleviating data scarcity in specialized scientific corpora, and there remain ample opportunities to build on this foundation across both technical and ethical dimensions.

\section{Acknowledgments}
This study is supported by grants from the Strategic Priority Research Program of the Chinese Academy of Sciences
XDB38030300 and Informatization
Plan of Chinese Academy of Sciences (CAS-WX2021SF-0101, CAS-WX2021SF-0111).
We would thank Dr. Ziyue Qiao for his valuable comments on this research. 

\bibliographystyle{IEEEtran}
\bibliography{ref}

\end{document}